\documentclass{article}
\pdfoutput=1
\usepackage{stfloats} 
\usepackage{framed}
\usepackage[most]{tcolorbox}
\usepackage{authblk}

\author[1]{Spyros Dragazis} \author[1]{Yilei Chen} \author[1,2]{Aldo Pacchiano} \affil[1]{Boston University} \affil[2]{Broad Institute of MIT and Harvard} 

\usepackage{graphicx} 
\usepackage{amsmath,amscd,amssymb,amsthm}
\usepackage{thmtools}
\usepackage{thm-restate}
\usepackage[margin=1in]{geometry}

\newtheorem{assum}{Assumption}
\newtheorem{lemma}{Lemma}

\usepackage[numbers,sort&compress]{natbib}

\usepackage{algorithm}      
\usepackage{algpseudocode}  
\usepackage{algorithmicx}   

\usepackage{enumitem}
\usepackage[dvipsnames]{xcolor}  
\usepackage{tcolorbox} 
\usepackage{manfnt} 
\usepackage{bbding}    

\usepackage[backref=page]{hyperref}
\hypersetup{
    colorlinks,
    citecolor=blue,
    filecolor=blue,
    linkcolor=blue,
    urlcolor=blue
}
\renewcommand*{\backref}[1]{}
\renewcommand*{\backrefalt}[4]{%
    \ifcase #1 %
    \or     #2%
    \else   #2%
    \fi
}

\usepackage{mymacros}

\usepackage{fontawesome5}

\usepackage{mathtools} 

\usepackage{manfnt} 

\usepackage{parskip}

\usepackage{pifont} 

\usepackage{ulem} 

\usepackage{nicefrac}

\usepackage{cleveref}

\crefname{proposition}{Proposition}{Propositions}
\Crefname{proposition}{Proposition}{Propositions}

\crefname{assum}{Assumption}{Assumptions}
\Crefname{assum}{Assumption}{Assumptions}

\crefname{thm}{Theorem}{Theorems}
\Crefname{thm}{Theorem}{Theorems}

\crefname{defn}{Definition}{Definitions}
\Crefname{defn}{Definition}{Definitions}

\usepackage{mdframed}

\usepackage{tikz}
\usetikzlibrary{arrows.meta,positioning,fit,backgrounds}

\usepackage{thm-restate} 

\usepackage{cancel}

\usepackage{color-edits}
\addauthor[Spyros]{sd}{NavyBlue}
\addauthor[Aldo]{ap}{ForestGreen}
\addauthor[Yilei]{yc}{SeaGreen}

\title{Explore Once, Evaluate Many: Multiple Policy Evaluation for Stochastic Linear Bandits}

\begin{document}

\maketitle

\begin{abstract}%
We study how to evaluate $K$ target policies simultaneously
in stochastic contextual bandits with linear reward structure.
Given a known $d$-dimensional feature map and reward-free
access to policy queries, we propose PRISMA (Policy evaluation through Regularized
Information Sharing and Mixture Allocation), an algorithm
that estimates every target policy's value to accuracy
$\varepsilon$ with probability at least $1-\delta$.
The algorithm uses
$\widetilde{\mathcal{O}}\!\left(
\frac{(L^2S^2+\beta^2)(1\vee\rho^\star)}{\varepsilon^2} \cdot \log(K)
\right)$ context draws and reward observations,
where $\rho^\star\le d$ is an instance-dependent measure
of the difficulty of designing a shared sampling distribution.
Here, $L$, $S$, and $\beta$ bound the feature norm,
reward parameter norm, and reward noise, respectively,
and the suppressed factors are logarithmic in
$K$, $d$, and $1/\delta$.
Our method combines reward-free planning, a regularized
sampling design, and a single ridge regression estimator
shared across all target policies.
An adaptive doubling procedure selects the regularization
scale without knowledge of $\rho^\star$.
The guarantees require no numerical lower bound on
second-moment eigenvalues and have no explicit dependence
on the cardinalities of the context and action spaces.
Policy-query costs are accounted for separately,
with $K$ queries per shared planning context.
\end{abstract}

\section{Introduction}\label{sec:introduction}

Contextual bandits provide a framework for sequential
decision-making in applications such as recommendation
systems \citep{li2010contextual,li2011unbiased},
medicine \citep{bastani2020online,trella2025deployed,brunskill2026active},
finance \citep{zhu2021online,soemers2018adapting},
and operations research
\citep{badanidiyuru2014resourceful,javanmard2019dynamic}.
Linear reward models are particularly useful because they
allow information to be shared across contexts and actions
through a feature representation.
A variety of learning approaches have been developed,
including optimistic algorithms for linear bandits
\citep{abbasi2011improved} and methods using neural
representations
\citep{zhou2020neural,bui2024variance,xu2020neural}.
Different training procedures, model architectures, and
hyperparameter choices can produce many candidate policies.
However, it is often unclear which of these policies
will perform best in the population where it will be deployed.

Consider a clinician choosing among several pretrained
models that recommend treatments from patient measurements.
Querying these models on existing patient records may be
inexpensive, whereas evaluating their recommendations
requires administering treatments and observing outcomes.
A similar question arises when a recommendation platform
enters a new market: a policy trained for one population
may perform differently in another, and several candidate
policies may need to be evaluated on the new population.
In both examples, the goal is to assess a collection of
fixed policies while limiting the cost of collecting
reward observations.

Estimating each policy's expected reward to a prescribed
accuracy supports decisions such as selecting a
near-optimal policy or identifying policies whose values
exceed a desired threshold.
A direct approach evaluates each candidate separately.
However, this requires a separate reward budget for every
policy, even when their decisions overlap or their feature
representations allow observations to be shared.
This motivates multiple-policy evaluation: estimating
the values of all target policies simultaneously with
high probability.

\citet{dann2023multiple} studied this problem in tabular
Markov decision processes, exploiting similarities among
target policies to reduce the cost of evaluation.
\citet{chen2024multiple} developed an approach based on
constructing a shared sampling distribution and estimating
policy values through importance weighting.
These works establish guarantees for tabular models.
Extending the underlying ideas to large context and action
spaces requires exploiting structure beyond individual
context--action pairs.
In particular, under linear reward structure, we would
like the evaluation cost to reflect the feature dimension
and the geometry of the target policies.

This leads to the following question:
\begin{quote}
\itshape
Can we evaluate a collection of target policies simultaneously
under linear reward structure, with only logarithmic explicit
dependence on the number of policies and no dependence on
the cardinalities of the context and action spaces in the
context and reward sample bounds?
\end{quote}

In this work, we answer this question affirmatively for
stochastic contextual bandits with a known feature map
and a linear conditional mean reward.
We consider $K$ fixed target policies and allow
reward-free queries to each policy on a shared context.
We propose PRISMA, which uses reward-free policy queries
on shared contexts to construct a sampling mixture and
then fits a single linear reward model to evaluate all
target policies.

With probability at least $1-\delta$, our algorithm
estimates every target policy's value to accuracy
$\varepsilon$ using
\[
    \widetilde{\CO}\left(
        \frac{(L^2S^2+\beta^2)(1\vee\rho^\star)}{\varepsilon^2} \cdot \log(K)
    \right)
\]
context draws and reward observations, where
$\widetilde{\CO}$ suppresses logarithmic factors in
$K$, $d$, and $1/\delta$.
Here, $L$ bounds the feature norm, $S$ bounds the reward
parameter norm, and $\beta$ bounds the reward noise.
The instance-dependent quantity $\rho\opt$ measures
the smallest worst-case value-estimation uncertainty
achievable by a mixture of the target policies,
and satisfies $\rho\opt\le d$.
We account separately for policy queries: each shared
planning context requires one query to every target policy.

A central challenge is to learn a useful sampling design
from estimated feature moments without requiring a
numerical lower bound on the eigenvalues of their
second-moment matrices.
We address this challenge through a regularized design
objective and multiplicative concentration of regularized
second moments.
An adaptive doubling procedure selects the regularization
scale and sample budgets without knowledge of $\rho\opt$.
Together, these ingredients yield an instance-dependent
evaluation guarantee while avoiding explicit dependence
on the sizes of the context and action spaces.
\footnote{Because of space constraints, we direct the reader to \Cref{sec:overview,sec:related_work} for an analytical overview of our approach and a discussion of related work.}

\section{Problem Formulation}\label{sec:problem_formulation}

\paragraph{Notation.}
For a positive integer $n$, let $[n] := \{1,\ldots,n\}$.
For a finite set $\mathcal S$, we denote its cardinality by
$|\mathcal S|$. We write $\triangle(\mathcal S)$ for the set of
probability distributions on a measurable space $\mathcal S$, and
$\triangle^{n-1} := \{p \in [0,1]^n : \sum_{i=1}^n p_i = 1\}$
for the probability simplex over $n$ elements.
For vectors $u,v$, let $\langle u,v\rangle := u^\top v$,
and let $\|u\|_2$ denote the Euclidean norm.
For symmetric matrices $A,B$ of the same size, we write
$A \preceq B$ if $B-A$ is positive semidefinite, and $A \succ 0$
if $A$ is positive definite.
For $A \succ 0$, define $\|u\|_A := \sqrt{u^\top A u}$.
We denote the identity matrix by $I$, with its dimension understood
from context.
The notation $\operatorname*{ess\,sup}$ denotes the essential
supremum with respect to the indicated probability distribution.
All logarithms are natural.
For nonnegative quantities $a,b$, we write $a \lesssim b$ if
$a \le Cb$ for a universal constant $C>0$.
The notation $\widetilde{\mathcal O}$ suppresses logarithmic factors.


We consider a stochastic contextual linear bandit \citep{krishnamurthy2023proportional,zanette2021design,brunskill2026active} with context space
$\mathcal X$, action space $\mathcal A$, and an unknown context
distribution $\mathcal P$.
The learner has sampling access to $\mathcal P$ and access to $K$
known target policies $\Pi^K := \{\pi_1,\ldots,\pi_K\}$, where each
policy $\pi_i:\mathcal X\to\triangle(\mathcal A)$ maps a context to a
distribution over actions.
Contexts are sampled independently from $\mathcal P$.
Given a context $X_t$, the learner can select a policy index
$I_t\in[K]$ and sample an action
$A_t\sim\pi_{I_t}(\cdot\mid X_t)$.
Executing this action produces a reward
\begin{equation}\label{eq:reward_mechanism}
    R_t=\langle\phi(X_t,A_t),\theta^\star\rangle+\eta_t,
\end{equation}
where $\phi:\mathcal X\times\mathcal A\to\mathbb R^d$ is a known
feature map and $\theta^\star\in\mathbb R^d$ is an unknown, fixed
reward parameter.
Conditional on $(X_t,A_t)$, the reward is drawn independently of
the past from a fixed distribution.
The noise satisfies
\[
    \mathbb E[\eta_t\mid X_t,A_t]=0,
    \qquad
    |\eta_t|\le\beta
    \quad\text{almost surely},
\]
for a known constant $\beta>0$.

For any sampled context $x$, the learner can query each target
policy and compute the feature vectors of the sampled actions
without executing those actions or observing their rewards.
A reward is observed only when an action is executed.
We distinguish between context samples, policy queries, and reward
observations when accounting for sample complexity.
The context and action spaces may be infinite; our goal is to
obtain guarantees expressed through the feature dimension $d$,
without dependence on the cardinalities of these spaces $|\CX \times \CA|$.

For any policy $\pi$, we define its value as
\begin{equation}\label{eq:value function}
    V_\pi
    := \E_\pi[R]
    = \E_\pi\left[\dotp{\phi(x,a)}{\theta\opt}\right],
\end{equation}
where $\E_\pi$ denotes expectation over $x\sim\CP$,
$a\sim\pi(\cdot\mid x)$, and the reward noise when applicable.
The equality follows from $\E[\eta\mid x,a]=0$.
In particular, for any integrable function $Y(x,a)$, we write
$\E_\pi Y(x,a)$ as shorthand for
$\E_{x\sim\CP}\E_{a\sim\pi(\cdot\mid x)}Y(x,a)$.

Given a set of target policies $\Pi^K$, sampling access to $\CP$,
and the interaction model described above, the task of
\textit{multiple-policy high-confidence policy evaluation}
\citep{dann2023multiple,chen2024multiple}
is to estimate the values of all target policies simultaneously.
Specifically, given an accuracy $\varepsilon>0$ and a confidence
parameter $\delta\in(0,1)$, the learner must output estimates
$\hat V_1,\ldots,\hat V_K$ satisfying
\begin{equation}\label{eq:policy-evaluation-guarantee}
    \P\left(
        \max_{i\in[K]}\abs{\hat V_i-V_i}\le\varepsilon
    \right)\ge 1-\delta,
\end{equation}
where $V_i:=V_{\pi_i}$.
The probability is over the randomness in data collection and
the algorithm.

Before presenting our algorithm and its sample complexity
guarantees, we state the following structural assumptions.

\begin{assum}[$\ell_2$-bounded features and parameter]
\label{assum:bounded_params}
    There exist known constants $L,S>0$ such that
    \begin{align*}
        \norm{\phi(x,a)}_2 &\le L,
        \qquad \forall (x,a)\in\CX\times\CA, \\
        \norm{\theta\opt}_2 &\le S.
    \end{align*}
\end{assum}

We use these bounds to control feature estimation errors and the
regularization bias in value estimation.
Knowledge of $L$ and $S$ is a standard assumption in classical
linear bandit algorithms, such as OFUL \citep{abbasi2011improved}.
Developing an algorithm that does not require these bounds is
beyond the scope of this work.
For linear bandit algorithms that do not require prior knowledge
of an upper bound on $\norm{\theta\opt}_2$, see
\citet{gales2022norm}.

\begin{assum}[Coverage]\label{assum:coverage}
    For every target policy $\pi_i$, $i\in[K]$, the feature
    second-moment matrix
    \[
        \Sigma_i
        := \E_{\pi_i}\left[\phi(x,a)\phi(x,a)^\top\right]
    \]
    is positive definite, i.e., $\Sigma_i\succ 0$.
\end{assum}

This assumption requires the feature distribution induced by each
target policy to span $\mathbb R^d$.
Similar full-rank conditions appear in policy evaluation with
linear function approximation; see, for example,
\citet{bhandari2018finite}.
We impose no quantitative lower bound on
$\lambda_{\min}(\Sigma_i)$ for any $i\in[K]$.

\section{Planning Phase}\label{sec:planning}

Designing a sampling distribution that allows a shared dataset to
evaluate all target policies is a natural approach to multiple-policy
evaluation. For tabular MDPs, \citet{chen2024multiple} use coarse
estimates of visitation distributions to construct an approximately
optimal sampling distribution, and then estimate importance weights
to evaluate the target policies. \citet{dann2023multiple} exploit
overlap between target policies by coupling their rollouts and sharing
samples. Explicitly representing visitation distributions and
importance weights requires quantities indexed by state--action pairs.
Such a representation can be costly when $|\CX\times\CA|$ is large,
and is unsuitable when these spaces are infinite. We therefore use the
known feature map to formulate the sampling design problem in terms
of the feature dimension $d$.

In a finite context--action space, the usual importance weight for
evaluating a target policy $\pi$ using samples from a behavior policy
is the target-to-behavior ratio
$\P_\pi(X=x,A=a)/\P_{\bmu}(X=x,A=a)$,
provided that the behavior distribution covers the target distribution.
In the linear setting, we can instead construct weights using feature
means and second moments.

We write
\[
    \bphi{\pi} := \E_\pi[\phi(X,A)]
\]
for the mean feature vector of policy $\pi$.
For a mixture $\bmu=(\mu_1,\ldots,\mu_K)\in\triangle^{K-1}$, let
\[
    \pi_{\bmu}(\cdot\mid x)
    := \sum_{i=1}^K \mu_i\pi_i(\cdot\mid x).
\]
We use $\E_{\bmu}$ to denote expectation under this mixture policy
and the common context distribution $\CP$. Its feature second-moment
matrix is
\begin{equation}\label{eq:sigma_mu_defn}
    \Sigma_{\bmu}
    := \E_{\bmu}\!\left[\phi(X,A)\phi(X,A)^\top\right]
    = \sum_{i=1}^K \mu_i\Sigma_i.
\end{equation}
By \Cref{assum:coverage}, $\Sigma_{\bmu}$ is positive definite for
every $\bmu\in\triangle^{K-1}$.

For any target policy $\pi$, we can therefore write
\begin{align}
    V_\pi
    &= \E_\pi\!\left[\dotp{\phi(X,A)}{\theta\opt}\right]
     = \bphi{\pi}^\top\theta\opt
     \notag\\
    &= \bphi{\pi}^\top\Sigma_{\bmu}^{-1}
       \E_{\bmu}\!\left[\phi(X,A)\phi(X,A)^\top\right]
       \theta\opt
     \notag\\
    &= \E_{\bmu}\!\left[
       \bphi{\pi}^\top\Sigma_{\bmu}^{-1}\phi(X,A)
       \dotp{\phi(X,A)}{\theta\opt}
       \right].
       \label{eq:change_of_measure_linear}
\end{align}
Thus, the appropriate analogue of an importance weight in our
linear setting is
\begin{equation}\label{eq:w(x,a)defn}
    w_\pi(x,a)
    := \bphi{\pi}^\top\Sigma_{\bmu}^{-1}\phi(x,a).
\end{equation}
These weights may be negative. Their role is to match the target
feature mean under the behavior distribution, which is sufficient
to recover the target value under the linear reward model.

To motivate the choice of mixture, first suppose that the population
quantities $\bphi{\pi}$ and $\Sigma_{\bmu}$ are known.
Fix $\bmu\in\triangle^{K-1}$ and consider a fresh batch of $N$
independent samples collected according to this mixture. For each
$t\in[N]$, we draw
\[
    X_t\sim\CP,\qquad
    I_t\sim\bmu,\qquad
    A_t\sim\pi_{I_t}(\cdot\mid X_t),
\]
where $I_t$ is independent of $X_t$, and observe a reward $R_t$
generated according to \Cref{eq:reward_mechanism}.
For any target policy $\pi$, define
\begin{equation}\label{eq:hatVpi}
    \hat{V}_\pi
    := \frac{1}{N}\sum_{t=1}^N
       w_\pi(X_t,A_t)R_t.
\end{equation}
By \Cref{eq:change_of_measure_linear} and the conditional
mean-zero noise assumption, this estimator is unbiased:
$\E[\hat{V}_\pi]=V_\pi$.
If $\bmu$ is chosen using independent planning data, the same
argument applies conditional on those data.

The second moment of the weight satisfies
\[
    \E_{\bmu}\!\left[w_{\pi_k}(X,A)^2\right]
    = \bphi{k}^\top\Sigma_{\bmu}^{-1}\bphi{k}.
\]
Since $|R|\le LS+\beta$ almost surely, it follows that
\[
    \operatorname{Var}_{\bmu}\!\left(w_{\pi_k}(X,A)R\right)
    \le (LS+\beta)^2
       \bphi{k}^\top\Sigma_{\bmu}^{-1}\bphi{k}.
\]
A Bernstein concentration argument therefore gives a
value-estimation bound whose leading variance term is controlled by
\[
    \rho(\bmu)
    := \max_{k\in[K]}
       \bphi{k}^\top\Sigma_{\bmu}^{-1}\bphi{k}.
\]
This motivates choosing a mixture that minimizes $\rho(\bmu)$,
and defining the instance-dependent quantity
\[
    \rho\opt
    := \min_{\bmu\in\triangle^{K-1}}\rho(\bmu).
\]
The full Bernstein bound also contains a term controlling the
maximum magnitude of the weighted rewards. Of course, the population
quantities $\bphi{k}$ and $\Sigma_k$ are unknown and must be estimated
during planning. We first state the concentration guarantee when
these quantities are known, and then explain how to estimate them.

\begin{restatable}{lemma}{Vhat_population}\label{lem:V_hat_population}
    Suppose that the population feature means $\bphi{k}$ and
    second moments $\Sigma_k$, $k\in[K]$, are known.
    Fix $\bmu\in\triangle^{K-1}$, and let
    $\{(X_t,A_t,R_t)\}_{t=1}^N$ be $N\ge 1$ independent samples
    collected under the mixture policy $\pi_{\bmu}$ and the reward
    model in \Cref{eq:reward_mechanism}.
    Define
    \[
        W_{\bmu}
        := \max_{k\in[K]}
        \operatorname*{ess\,sup}_{X\sim\CP,\,
            A\sim\pi_{\bmu}(\cdot\mid X)}
        \abs{w_{\pi_k}(X,A)}.
    \]
    For any $\delta\in(0,1)$, with probability at least $1-\delta$,
    the estimators in \Cref{eq:hatVpi} satisfy
    \[
        \max_{k\in[K]}
        \abs{\hat{V}_{\pi_k}-V_{\pi_k}}
        \le (LS+\beta)
        \left(
            \sqrt{\frac{2\rho(\bmu)\log(2K/\delta)}{N}}
            + \frac{4W_{\bmu}\log(2K/\delta)}{3N}
        \right).
    \]
    If the mixture is selected using independent planning data,
    the same guarantee holds conditional on those data.
\end{restatable}

Consequently, to estimate all target policy values with
$\varepsilon$ accuracy, it is sufficient to take
\[
    N \ge \log\!\left(\frac{2K}{\delta}\right)
    \max\left\{
        \frac{8(LS+\beta)^2\rho(\bmu)}{\varepsilon^2},
        \frac{8(LS+\beta)W_{\bmu}}{3\varepsilon}
    \right\}.
\]
The leading variance term therefore scales as
$\tilde{\CO}((LS+\beta)^2\rho(\bmu)/\varepsilon^2)$.
The population benchmark $\rho\opt$ minimizes this uncertainty
over all mixtures of the target policies.
Since the feature means $\bphi{k}$ and second moments $\Sigma_k$
are unknown, we will estimate them and use a regularized empirical
design problem during planning.
We first show that the population benchmark is at most $d$.

\begin{restatable}{proposition}{maxUncertaintyBound}
\label{prop:bound_max_uncertainty}
    Under \Cref{assum:coverage}, the optimal design complexity satisfies
    \[
        \rho\opt
        = \min_{\bmu\in\triangle^{K-1}}
          \max_{k\in[K]}
          \bphi{k}^\top
          \left(\sum_{i=1}^K\mu_i\Sigma_i\right)^{-1}
          \bphi{k}
        \le d.
    \]
\end{restatable}

In \citet{chen2024multiple}, coarse estimates of policy visitation
distributions are used to construct a sampling distribution for
multiple-policy evaluation in tabular MDPs.
Coarse estimation allows the cost of estimating visitation
distributions to scale with $1/\varepsilon$, up to logarithmic
factors and factors depending on the tabular problem parameters,
rather than $1/\varepsilon^2$.
In our setting, we do not necessarily need to apply the coarse estimation technique as the context distribution does not depend on the
policy, and we can query every target policy on the same context
without observing rewards.
Thus, for each context $X_t \sim \CP$, we sample
$A_t^k \sim \pi_k(\cdot \mid X_t)$ and compute
$\phi(X_t,A_t^k)$ for every $k \in [K]$.
This allows us to estimate the feature means and second moments
directly from a shared batch of contexts.


We begin by estimating the unknown population quantities
$\bphi{k}$ and $\Sigma_k$.
Draw $X_1,\ldots,X_N$ independently from $\CP$ and, for each
$t \in [N]$, sample $A_t^k \sim \pi_k(\cdot \mid X_t)$
using fresh randomness.
For each fixed policy $k$, the feature vectors
$\{\phi(X_t,A_t^k)\}_{t=1}^N$ are therefore independent
and identically distributed, with mean $\bphi{k}$ and
second moment $\Sigma_k$.
We finally obtain simultaneous guarantees through
a union bound.

Since the feature vectors satisfy
$\|\phi(X_t,A_t^k)\|_2 \le L$ by
\Cref{assum:bounded_params}, we first use McDiarmid's inequality,
together with a bound on the expected estimation error,
to control the Euclidean distance between each empirical
feature mean and its population counterpart.

\begin{restatable}{lemma}{hatPhiEstimation}
\label{lem:hat_phi_estimation}
    Let $X_1,\ldots,X_N$ be independent contexts drawn from $\CP$.
    For each $t \in [N]$ and $k \in [K]$, sample
    $A_t^k \sim \pi_k(\cdot \mid X_t)$ using fresh randomness,
    and define
    \[
        \hat{\phi}^k_N
        := \frac{1}{N}\sum_{t=1}^N \phi(X_t,A_t^k).
    \]
    Under \Cref{assum:bounded_params}, for any
    $\delta \in (0,1)$, with probability at least $1-\delta$,
    simultaneously for all $k \in [K]$,
    \[
        \norm{\hat{\phi}^k_N - \bphi{k}}_2
        \le
        \frac{L}{\sqrt{N}}
        \left(1+\sqrt{2\log\left(\frac{K}{\delta}\right)}\right).
    \]
\end{restatable}

See \Cref{app:proof_hat_phi_estimation} for the proof.

We next estimate the second moments used in the design problem.
Directly controlling the relative error of an estimate of
$\Sigma_k^{-1}$ generally requires a sample size that depends
on the smallest eigenvalue of $\Sigma_k$, which can be
arbitrarily small.
To avoid this dependence, we introduce a regularization
parameter $\lambda>0$ and estimate $\Sigma_k+\lambda I$
to constant multiplicative accuracy.
Since $\Sigma_k+\lambda I \succeq \lambda I$, the resulting
sample bound depends on $\lambda$ instead of a lower bound
on the eigenvalues of $\Sigma_k$.
We will choose $\lambda$ later to balance the cost of planning
with the regularization bias in value estimation.

\begin{restatable}{lemma}{hatSigmaEstimation}
\label{lem:hatSigmaEstimation}
    Let $X_1,\ldots,X_N$ be independent contexts drawn from $\CP$.
    For each $t \in [N]$ and $k \in [K]$, sample
    $A_t^k \sim \pi_k(\cdot \mid X_t)$ using fresh randomness,
    and define
    \[
        \hat{\Sigma}_k
        :=
        \frac{1}{N}\sum_{t=1}^N
        \phi(X_t,A_t^k)\phi(X_t,A_t^k)^\top.
    \]
    Under \Cref{assum:bounded_params}, fix $\lambda>0$
    and $\delta \in (0,1)$.
    If
    \[
        N \ge
        \frac{28L^2}{3\lambda}
        \log\left(\frac{2Kd}{\delta}\right),
    \]
    then, with probability at least $1-\delta$,
    simultaneously for all $k \in [K]$,
    \[
        \frac12(\Sigma_k+\lambda I)
        \preceq
        \hat{\Sigma}_k+\lambda I
        \preceq
        \frac32(\Sigma_k+\lambda I).
    \]
\end{restatable}

See \Cref{app:proof_hatSigmaEstimation} for the proof.
Since inversion reverses the positive definite order,
on the same event we also have
\[
    \frac23(\Sigma_k+\lambda I)^{-1}
    \preceq
    (\hat{\Sigma}_k+\lambda I)^{-1}
    \preceq
    2(\Sigma_k+\lambda I)^{-1},
    \qquad k \in [K].
\]

We now replace the population quantities by their empirical
estimates and solve the regularized design problem.
For convenience, write $\hat{\phi}_k:=\hat{\phi}_N^k$ and define
\[
    \hat{F}_\lambda(\bmu)
    :=
    \max_{k \in [K]}
    \hat{\phi}_k^\top
    \left(
        \sum_{i=1}^K \mu_i\hat{\Sigma}_i+\lambda I
    \right)^{-1}
    \hat{\phi}_k.
\]
We select a mixture
\[
    \hat{\bmu}
    \in
    \operatorname*{arg\,min}_{\bmu \in \triangle^{K-1}}
    \hat{F}_\lambda(\bmu).
\]
This is a convex optimization problem.
Indeed, by the Schur complement, it is equivalent to the
semidefinite program
\[
\begin{aligned}
    &\underset{\bmu \in \triangle^{K-1},\,z \in \mathbb{R}}
    {\operatorname{minimize}}
    && z \\
    &\text{subject to}
    &&
    \begin{pmatrix}
        \sum_{i=1}^K \mu_i\hat{\Sigma}_i+\lambda I
        & \hat{\phi}_k \\
        \hat{\phi}_k^\top & z
    \end{pmatrix}
    \succeq 0,
    \qquad k \in [K].
\end{aligned}
\]
Since $\lambda>0$, the objective is continuous on the
compact simplex, so a minimizer exists \citep{vandenberghe2004convex}.

To assess the selected mixture, define the corresponding
regularized population objective
\[
    F_\lambda(\bmu)
    :=
    \max_{k \in [K]}
    \bphi{k}^\top
    \left(
        \sum_{i=1}^K \mu_i\Sigma_i+\lambda I
    \right)^{-1}
    \bphi{k},
\]
and recall that
\[
    \rho\opt
    :=
    \min_{\bmu \in \triangle^{K-1}}
    \max_{k \in [K]}
    \bphi{k}^\top\Sigma_{\bmu}^{-1}\bphi{k}.
\]
The following lemma compares the empirical and population
objectives and bounds the uncertainty of the selected mixture
in terms of the unregularized benchmark $\rho\opt$.

\begin{restatable}{lemma}{sampleOptProblem}
\label{lem:sampleOptProblem}
    Under \Cref{assum:bounded_params,assum:coverage},
    fix $\lambda>0$ and $\delta \in (0,1)$.
    Form $\hat{\phi}_k$ and $\hat{\Sigma}_k$ from the same
    batch of $N$ independent contexts and fresh policy queries,
    as in \Cref{lem:hat_phi_estimation,lem:hatSigmaEstimation}.
    If
    \[
        N \ge
        \frac{16L^2}{\lambda}
        \log\left(\frac{4Kd}{\delta}\right),
    \]
    then, with probability at least $1-\delta$, every
    minimizer
    $\hat{\bmu}\in
    \operatorname*{arg\,min}_{\bmu\in\triangle^{K-1}}
    \hat{F}_\lambda(\bmu)$ satisfies
    \[
        \hat{F}_\lambda(\hat{\bmu})
        \le 4\rho\opt+4,
        \qquad
        F_\lambda(\hat{\bmu})
        \le 12\rho\opt+14.
    \]
\end{restatable}

See \Cref{app:proof_sampleOptProblem} for the proof.
We deploy this mixture in the sampling phase and account
for the resulting regularization bias when estimating
the policy values.

\section{Sampling phase}\label{sec:sampling_phase}

During the planning phase, we used reward-free feature samples
to select a mixture $\hat{\bmu}$ of the target policies.
We now deploy this mixture, collect reward observations,
and use them to estimate the values of all target policies.
The planning guarantees depend on a regularization parameter
$\lambda>0$.
In this section, we analyze how this parameter affects the
final estimation error and choose it, together with the sample
budgets, to achieve accuracy $\varepsilon$.

Our estimator combines feature-mean estimates with a single
linear reward model fitted by ridge regression.
For each target policy $\pi_k$, we estimate
$V_{\pi_k}=\bphi{k}^\top\theta\opt$ by
\[
    \hat{V}_k:=\tilde{\phi}_k^\top\hat{\theta},
\]
where $\tilde{\phi}_k$ is a fresh estimate of $\bphi{k}$
and $\hat{\theta}$ is fitted using rewards collected under
$\hat{\bmu}$.

To simplify the dependence structure in the analysis, we use
separate batches for estimating the feature means and fitting
the reward model.
Specifically, draw $N_{\mathrm{mean}}$ fresh contexts
$X'_t\sim\CP$, independently of the planning data and the
reward batch.
For each context, query every target policy to obtain
$A_t^{\prime k}\sim\pi_k(\cdot\mid X'_t)$, and define
\[
    \tilde{\phi}_k
    :=
    \frac{1}{N_{\mathrm{mean}}}
    \sum_{t=1}^{N_{\mathrm{mean}}}
    \phi(X'_t,A_t^{\prime k}),
    \qquad k\in[K].
\]
By \Cref{lem:hat_phi_estimation}, a batch of
\[
    N_{\mathrm{mean}}
    =
    \left\lceil
    \frac{L^2}{\lambda}
    \left(
        1+\sqrt{2\log\left(\frac{2K}{\delta}\right)}
    \right)^2
    \right\rceil
\]
contexts suffices to ensure
$\|\tilde{\phi}_k-\bphi{k}\|_2\le\sqrt{\lambda}$
simultaneously for all $k\in[K]$, with probability at least
$1-\delta/2$.
These contexts are shared across policies, so this step
requires $KN_{\mathrm{mean}}$ policy queries and no reward
observations.

We next collect $M$ fresh reward samples using the selected
mixture.
For each $i\in[M]$, draw $X_i\sim\CP$, independently sample
$I_i\sim\hat{\bmu}$, and execute
$A_i\sim\pi_{I_i}(\cdot\mid X_i)$.
Observe the reward $R_i$ generated according to
\Cref{eq:reward_mechanism}.
Conditional on the planning data, these samples are
independent and identically distributed under the fixed
mixture $\hat{\bmu}$.
Define
\begin{equation}\label{eq:ridge_quants}
\begin{aligned}
    \tilde{\Sigma}
    &:=
    \frac{1}{M}\sum_{i=1}^M
    \phi(X_i,A_i)\phi(X_i,A_i)^\top,\\
    \tilde{g}
    &:=
    \frac{1}{M}\sum_{i=1}^M
    \phi(X_i,A_i)R_i,\\
    B&:=\tilde{\Sigma}+\lambda I.
\end{aligned}
\end{equation}
Our ridge regression estimator is
$\hat{\theta}:=B^{-1}\tilde{g}$.


We first express the parameter-estimation error as the sum of
a regularization term and a reward-noise term.
This identity will yield a three-term decomposition of the
value-estimation error and guide our choice of $\lambda$
and the reward sample budget $M$.

\begin{restatable}{lemma}{thetaError}\label{lem:thetaError}
    Let $B$ and $\tilde{g}$ be defined in
    \Cref{eq:ridge_quants}, and let
    $\hat{\theta}:=B^{-1}\tilde{g}$.
    Under the reward model in \Cref{eq:reward_mechanism},
    \[
        \hat{\theta}-\theta\opt
        =
        -\lambda B^{-1}\theta\opt
        +
        \frac{1}{M}B^{-1}
        \sum_{i=1}^M \eta_i\phi(X_i,A_i).
    \]
\end{restatable}

See \Cref{app:proof_thetaError} for the proof.
This identity holds for every realization of the samples.

For brevity, write $\phi_i:=\phi(X_i,A_i)$.
Using \Cref{lem:thetaError}, we decompose the
value-estimation error as
\begin{equation}\label{eq:valueDecomp}
\begin{aligned}
    V_k-\hat{V}_k
    &=
    \bphi{k}^\top\theta\opt
    -\tilde{\phi}_k^\top\hat{\theta} \\
    &=
    (\bphi{k}-\tilde{\phi}_k)^\top\theta\opt
    +\tilde{\phi}_k^\top(\theta\opt-\hat{\theta}) \\
    &=
    \underbrace{
        (\bphi{k}-\tilde{\phi}_k)^\top\theta\opt
    }_{\text{feature-mean error}}
    +
    \underbrace{
        \lambda\tilde{\phi}_k^\top B^{-1}\theta\opt
    }_{\text{regularization bias}}
    -
    \underbrace{
        \frac{1}{M}\tilde{\phi}_k^\top B^{-1}
        \sum_{i=1}^M\eta_i\phi_i
    }_{\text{reward noise}}.
\end{aligned}
\end{equation}
We bound these three contributions separately, allocating
error $\varepsilon/2$ to feature-mean estimation and
$\varepsilon/4$ to each of the remaining terms.

For the feature-mean error, Cauchy--Schwarz and
\Cref{assum:bounded_params} give, on the event
$\|\tilde{\phi}_k-\bphi{k}\|_2\le\sqrt{\lambda}$,
\begin{align*}
    \left|
    (\bphi{k}-\tilde{\phi}_k)^\top\theta\opt
    \right|
    &\le
    \|\bphi{k}-\tilde{\phi}_k\|_2
    \|\theta\opt\|_2 \\
    &\le S\sqrt{\lambda}.
\end{align*}
Thus, this contribution is at most $\varepsilon/2$ whenever
\[
    \lambda\le\frac{\varepsilon^2}{4S^2}.
\]

To control the regularization bias, define
\[
    G:=\Sigma_{\hat{\bmu}}+\lambda I,
    \qquad
    Q_k:=\tilde{\phi}_k^\top B^{-1}\tilde{\phi}_k.
\]
Conditional on the planning data, the reward batch is drawn
from the fixed mixture $\hat{\bmu}$.
Applying the regularized second-moment concentration argument
of \Cref{lem:hatSigmaEstimation} to this mixture gives
\[
    \frac12G\preceq B\preceq\frac32G.
\]
We work on this event together with the planning and
feature-mean concentration events.

Writing $e_k:=\tilde{\phi}_k-\bphi{k}$, we have
$e_k^\top G^{-1}e_k\le\|e_k\|_2^2/\lambda\le1$.
Therefore,
\begin{align*}
    Q_k
    &\le
    2\tilde{\phi}_k^\top G^{-1}\tilde{\phi}_k \\
    &=
    2(\bphi{k}+e_k)^\top G^{-1}(\bphi{k}+e_k) \\
    &\le
    4\bphi{k}^\top G^{-1}\bphi{k}
    +4e_k^\top G^{-1}e_k \\
    &\le
    4F_\lambda(\hat{\bmu})+4 \\
    &\le 48\rho\opt+60,
\end{align*}
where the last step follows from
\Cref{lem:sampleOptProblem}.

For a positive definite matrix $H$, write
$\|v\|_H:=\sqrt{v^\top Hv}$.
Since $B\succeq\lambda I$, Cauchy--Schwarz in the
$B^{-1}$ inner product yields
\begin{align*}
    \lambda
    \left|\tilde{\phi}_k^\top B^{-1}\theta\opt\right|
    &\le
    \lambda
    \|\tilde{\phi}_k\|_{B^{-1}}
    \|\theta\opt\|_{B^{-1}} \\
    &\le
    \lambda\sqrt{Q_k}\,
    \frac{\|\theta\opt\|_2}{\sqrt{\lambda}} \\
    &\le
    S\sqrt{\lambda(48\rho\opt+60)}.
\end{align*}
Thus, the regularization bias is at most $\varepsilon/4$
provided that
\[
    \lambda
    \le
    \frac{\varepsilon^2}
    {16S^2(48\rho\opt+60)}.
\]
This condition also ensures that the feature-mean error
is at most $\varepsilon/2$.

It remains to control the reward-noise contribution.
Conditional on the planning data, the fresh feature-mean
batch, and the contexts and actions in the reward batch,
the coefficients
\[
    c_{i,k}:=
    \frac{1}{M}\tilde{\phi}_k^\top B^{-1}\phi_i
\]
are fixed.
The noise term is therefore a weighted sum
$\sum_{i=1}^M c_{i,k}\eta_i$.
The following lemma bounds these sums simultaneously
over the target policies.


\begin{restatable}{lemma}{boundNoiseTerm}
\label{lem:boundNoiseTerm}
    Let $B$ be defined in \Cref{eq:ridge_quants}, and let
    $\tilde{\phi}_k$ be the feature-mean estimates obtained
    from the independent batch.
    Define
    \[
        Q_k:=\tilde{\phi}_k^\top B^{-1}\tilde{\phi}_k.
    \]
    For any $\zeta\in(0,1)$, conditional on the planning data,
    the feature-mean batch, and the reward-batch contexts
    and actions, with probability at least $1-\zeta$,
    simultaneously for all $k\in[K]$,
    \[
        \left|
        \frac{1}{M}\tilde{\phi}_k^\top B^{-1}
        \sum_{i=1}^M\eta_i\phi(X_i,A_i)
        \right|
        \le
        \beta\sqrt{
        \frac{2Q_k\log(2K/\zeta)}{M}
        }.
    \]
    The same probability guarantee holds unconditionally.
\end{restatable}

See \Cref{app:proof_boundNoiseTerm} for the proof.
On the concentration events established above,
$Q_k\le48\rho\opt+60$ for every $k$.
Consequently, on their intersection with the noise event,
the reward-noise contribution is at most $\varepsilon/4$
whenever
\[
    M\ge
    \frac{32\beta^2(48\rho\opt+60)}{\varepsilon^2}
    \log\left(\frac{2K}{\zeta}\right).
\]
The reward sample budget must also be large enough to ensure
that $B$ concentrates around
$\Sigma_{\hat{\bmu}}+\lambda I$.
In particular, the bound on $Q_k$ used above requires
\[
    B \succeq \frac12(\Sigma_{\hat{\bmu}}+\lambda I).
\]
Conditional on the planning data, applying
\Cref{lem:hatSigmaEstimation} to the single selected mixture
shows that it suffices to require
\[
    M\ge
    \frac{28L^2}{3\lambda}
    \log\left(\frac{2d}{\zeta}\right).
\]

\section{Algorithm and guarantees}\label{sec:algorithm}

The preceding analysis identifies sufficient choices of the
regularization parameter and sample budgets in terms of
$\rho\opt$.
However, this quantity depends on unknown population moments.
We address this issue by testing increasing scales
$r_j=2^j$ and using the empirical design objective to decide
when a scale is sufficiently large.

At each stage, we collect a fresh planning batch and solve
the corresponding empirical design problem.
We accept the resulting mixture when
$\hat{F}_{\lambda_j}(\hat{\bmu}_j)\le9r_j$.
On the planning concentration event,
\Cref{lem:sampleOptProblem} gives
\[
    \hat{F}_{\lambda_j}(\hat{\bmu}_j)
    \le4\rho\opt+4.
\]
Consequently, a stage is accepted whenever
$r_j\ge1\vee\rho\opt$.
Since $\rho\opt\le d$ by
\Cref{prop:bound_max_uncertainty}, it suffices to consider
$j=0,\ldots,J$, where $J:=\lceil\log_2 d\rceil$.

After accepting a mixture, we collect the independent
feature-mean and reward batches described in
\Cref{sec:sampling_phase}.
To specify the budgets, let
\[
    \zeta:=\frac{\delta}{J+4},
    \qquad
    \ell:=\log\left(\frac{4Kd}{\zeta}\right).
\]
For each stage $j$, define
\begin{equation}\label{eq:stage_budgets}
\begin{aligned}
    r_j&:=2^j,
    &
    \lambda_j&:=\frac{\varepsilon^2}{2048S^2r_j},\\
    n_j&:=
    \left\lceil\frac{16L^2\ell}{\lambda_j}\right\rceil,
    &
    M_j&:=
    \left\lceil
    \max\left\{
        \frac{16L^2\ell}{\lambda_j},
        \frac{4096\beta^2r_j\ell}{\varepsilon^2}
    \right\}
    \right\rceil.
\end{aligned}
\end{equation}
We assume $L,S>0$ and $0<\varepsilon<LS$.
If $\varepsilon\ge LS$, returning zero for every policy
already achieves the desired accuracy.

\begin{algorithm}[t]
\caption{\texttt{PRISMA}: Policy evaluation through Regularized
Information Sharing and Mixture Allocation}
\label{alg:multiple_policy_evaluation}
\textbf{Input:}
Target policies $\{\pi_k\}_{k=1}^K$,
accuracy $\varepsilon$, confidence $\delta$,
and bounds $L,S,\beta$.

\begin{enumerate}
    \item Set $J=\lceil\log_2d\rceil$ and define the
    budgets in \Cref{eq:stage_budgets}.

    \item For $j=0,\ldots,J$:
    \begin{enumerate}
        \item Draw $n_j$ fresh contexts.
        On each context, query every target policy
        and compute its proposed action's feature vector.

        \item Using this batch, form
        $\hat{\phi}_{j,k}$ and $\hat{\Sigma}_{j,k}$
        as in \Cref{lem:hat_phi_estimation,lem:hatSigmaEstimation}.

        \item Form $\hat{F}_{\lambda_j}$ using these
        stage-specific estimates and compute
        \[
            \hat{\bmu}_j\in
            \operatorname*{arg\,min}_{\bmu\in\triangle^{K-1}}
            \hat{F}_{\lambda_j}(\bmu).
        \]

        \item If
        $\hat{F}_{\lambda_j}(\hat{\bmu}_j)\le9r_j$,
        set $\hat{j}=j$ and exit the loop.
    \end{enumerate}

    \item Set
    $\hat{\bmu}=\hat{\bmu}_{\hat{j}}$,
    $\lambda=\lambda_{\hat{j}}$,
    and $M=M_{\hat{j}}$.

    \item Use $n_{\hat{j}}$ fresh shared contexts and
    queries to all target policies to form
    $\tilde{\phi}_1,\ldots,\tilde{\phi}_K$,
    independently of the planning and reward batches.

    \item Collect $M$ fresh reward observations under
    $\hat{\bmu}$.
    Form $B$ and $\tilde{g}$ as in \Cref{eq:ridge_quants},
    and set $\hat{\theta}=B^{-1}\tilde{g}$.

    \item Return
    $\hat{V}_k=\tilde{\phi}_k^\top\hat{\theta}$
    for every $k\in[K]$.
\end{enumerate}
\end{algorithm}

The acceptance rule also controls the uncertainty of the
selected mixture.
Indeed, the comparisons established in the preceding
analysis give, on the concentration events,
\begin{align*}
    Q_k
    &\le4F_{\lambda_{\hat{j}}}(\hat{\bmu})+4\\
    &\le12\hat{F}_{\lambda_{\hat{j}}}(\hat{\bmu})+12\\
    &\le108r_{\hat{j}}+12
    \le120r_{\hat{j}}.
\end{align*}
The choices in \Cref{eq:stage_budgets} therefore control
both regularization bias and reward noise using the
accepted scale.

\begin{restatable}{theorem}{mainGuarantee}
\label{thm:main_guarantee}
    Suppose \Cref{assum:bounded_params,assum:coverage}
    and the stochastic reward model in
    \Cref{eq:reward_mechanism} hold.
    Let $0<\varepsilon<LS$ and $\delta\in(0,1)$.
    Assume the empirical design problems in
    \texttt{PRISMA} (\Cref{alg:multiple_policy_evaluation}) are solved exactly.

    \texttt{PRISMA}
    terminates by stage $J=\lceil\log_2 d\rceil$.
    With probability at least $1-\delta$,
    \[
        \max_{k\in[K]}|\hat{V}_k-V_k|\le\varepsilon,
        \qquad
        r_{\hat{j}}\le2(1\vee\rho\opt).
    \]
    On the same event, the total number of context draws
    and the number of reward observations are each at most
    \[
        \CO\left(
            \frac{(L^2S^2+\beta^2)(1\vee\rho\opt)}
                 {\varepsilon^2}
            \log\left(\frac{Kd\log(d+4)}{\delta}\right)
        \right).
    \]
    The constants hidden in $\CO(\cdot)$ are universal.
    For every realization, the corresponding resource
    bounds hold with $1\vee\rho\opt$ replaced by $d$.
\end{restatable}

See \Cref{app:proof_main_guarantee} for the proof.

\section{Discussion}\label{sec:discussion}

We studied simultaneous policy evaluation in stochastic
contextual bandits with linear reward structure.
Our results show that reward-free planning and a shared
sampling design yield instance-dependent guarantees
without explicit dependence on the cardinalities of
the context and action spaces or a numerical lower
bound on second-moment eigenvalues.

A natural direction is to extend these techniques to
linear MDPs and more general function approximation settings.
In these settings, an additional challenge is that
the policies influence the distributions of future states,
so feature moments cannot be estimated directly from
a shared policy-independent context distribution.

An open question is whether $\rho\opt$ captures the
intrinsic difficulty of evaluation in the linear setting.
Our analysis provides an upper bound in terms of this
quantity, but does not establish its necessity.
Matching lower bounds would help determine when mixtures
of target policies achieve the optimal evaluation cost
and whether more general sampling strategies can improve
upon them.

\section*{AI disclosure}
We used GPT-6 Astra and Sol through iterative, back-and-forth
discussions to polish the manuscript, work through calculations
in the concentration arguments, and check mathematical details.
These discussions also helped identify gaps and refine the
presentation of proofs.
The authors are responsible for the correctness and originality
of the final manuscript.


\nocite{*}
\bibliographystyle{plainnat}
\bibliography{refs}

\appendix

\crefalias{section}{appendix} 

  \hsize\textwidth
  \linewidth\hsize  {\centering
  {\Large\bfseries Appendix \par}}
  \vskip 0.2in
\tableofcontents
 \addcontentsline{toc}{section}{Supplementary Material}
\newpage
\section{Overview of our method}\label{sec:overview}

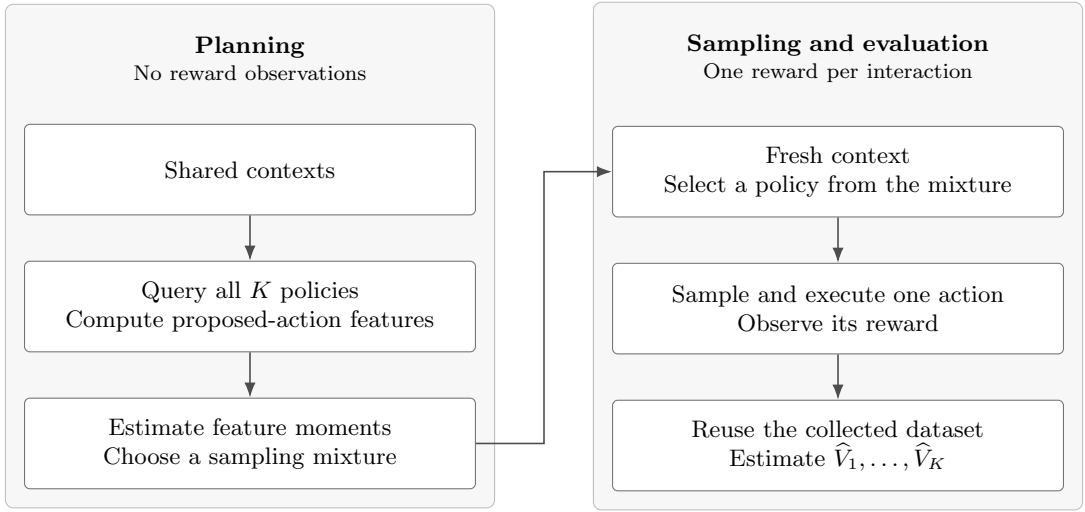
\begin{figure}[t]
  \centering
  \begin{tikzpicture}[
    font=\small,
    box/.style={
      draw=black!55, rounded corners=2pt, fill=white,
      text width=0.34\linewidth, align=center,
      minimum height=12mm, inner sep=5pt
    },
    heading/.style={
      text width=0.34\linewidth, align=center,
      inner sep=5pt
    },
    flow/.style={
      -{Latex[length=2mm]}, semithick, draw=black!70
    },
    panel/.style={
      draw=black!25, rounded corners=3pt,
      fill=black!3, inner sep=7pt
    }
  ]

    \node[heading] (plan-title)
      {\textbf{Planning}\\[-1pt]
       {\footnotesize No reward observations}};

    \node[box, below=4mm of plan-title] (contexts)
      {Shared contexts};

    \node[box, below=6mm of contexts] (proposals)
      {Query all $K$ policies\\
       Compute proposed-action features};

    \node[box, below=6mm of proposals] (mixture)
      {Estimate feature moments\\
        Choose a sampling mixture};

    \node[heading, right=18mm of plan-title] (sample-title)
      {\textbf{Sampling and evaluation}\\[-1pt]
       {\footnotesize One reward per interaction}};

    \node[box, below=4mm of sample-title] (selection)
      {Fresh context\\
       Select a policy from the mixture};

    \node[box, below=6mm of selection] (reward)
      {Sample and execute one action\\
       Observe its reward};

    \node[box, below=6mm of reward] (values)
      {Reuse the collected dataset\\
       Estimate $\widehat V_1,\ldots,\widehat V_K$};

    \draw[flow] (contexts) -- (proposals);
    \draw[flow] (proposals) -- (mixture);
    \draw[flow] (selection) -- (reward);
    \draw[flow] (reward) -- (values);
    \draw[flow] (mixture.east)
      -- ++(9mm,0) |- (selection.west);

    \begin{scope}[on background layer]
      \node[panel,
        fit=(plan-title)(contexts)(proposals)(mixture)] {};
      \node[panel,
        fit=(sample-title)(selection)(reward)(values)] {};
    \end{scope}

  \end{tikzpicture}
  \caption{Overview of the two-phase procedure. Planning queries all
  target policies on shared contexts without observing rewards. Sampling
  executes one action per context, and the resulting reward dataset is
  reused to estimate all $K$ policy values.
  Final value estimates also use feature means estimated from
a separate, fresh reward-free batch of contexts.}
  \label{fig:method-overview}
\end{figure}

Our method consists of two phases: \textit{planning} and \textit{sampling},
illustrated in Figure~\ref{fig:method-overview}.
During planning, we use contexts and queries to the target policies
to estimate their feature means and second moments, without observing
rewards. These estimates define a regularized design problem that
selects a mixture of target policies to control the largest
value-estimation uncertainty across them. The planning procedure also
determines the regularization parameter and the sample budgets used
in the next phase.

During sampling, we deploy the selected mixture: for each fresh context,
we sample a target policy and execute an action drawn from that policy.
We use the resulting reward observations to fit a single linear reward
model by ridge regression. Combining this model with target feature
means estimated from a separate, fresh batch of contexts yields value
estimates for all $K$ policies. A related separation between planning
and sampling appears in \citet{zanette2021design}, for the objective
of learning a near-optimal policy.

The distinction between context samples, policy queries, and reward
observations is central to our approach. For each context $x$ used
to estimate feature moments, we sample an action
$a_i\sim\pi_i(\cdot\mid x)$ from every target policy and compute
$\phi(x,a_i)$. This requires $K$ policy queries, but only one context
draw and no reward observations. Thus, policy-queries per context scale
linearly with $K$, while the same contexts are shared across all
target policies. Our primary objective is to reduce the number of
reward observations, and we account for these resources separately.

This distinction is also practically relevant. For example, contexts
may represent patient measurements and target policies may be
pretrained models recommending treatments. Querying these models on
historical patient records can be inexpensive, whereas observing a
reward requires administering a treatment and measuring its outcome.
An independent historical dataset whose contexts are drawn i.i.d.\
from the same distribution $\CP$ can therefore supply the planning
contexts.

\section{Related work}\label{sec:related_work}

\citet{dann2023multiple} studied multiple-policy
high-confidence evaluation in finite-horizon tabular MDPs.
Their goal, as in our work, is to estimate the values of
all target policies to a prescribed accuracy while sharing
observations across policies.
They achieve this through coupled trajectory sampling:
when policies choose the same action at a visited state,
a transition sample can be reused in their respective
trajectories.
Their guarantees depend on an expected-disagreement
quantity that captures the overlap among target policies.
They first analyze this procedure under a sampling oracle
and then construct an approximate oracle through interaction
with the environment.
Our approach exploits a different source of shared
information: under linear reward structure, an observation
can inform the values of policies taking different actions
through their feature representations.
We capture this relationship through policy feature means
and second moments.

The work closest to ours is \citet{chen2024multiple},
who propose CAESAR for multiple-policy evaluation in
tabular MDPs.
Their method first obtains coarse estimates of policy
visitation distributions and uses them to construct a
shared sampling distribution.
It then estimates marginal importance ratios through
quadratic objectives inspired by DualDICE and uses these
ratios to evaluate the target policies.
Their sample guarantee depends on an optimized
occupancy-based measure of the difficulty of covering
the target policies.
Our method follows a related separation between designing
a sampling distribution and collecting data for evaluation.
In our setting, the design is expressed through feature
moments, and its quality is measured by the largest
quadratic uncertainty among the target feature means.
Moreover, contexts follow a policy-independent distribution,
so we can estimate these moments by querying all policies
on shared contexts without observing rewards.
We use a regularized empirical design and a single ridge
regression model to obtain value estimates, accounting
explicitly for moment-estimation error and regularization
bias.

\citet{liu2025efficient} also design a shared behavior
policy to evaluate multiple target policies.
Their approach uses per-decision importance sampling
and derives behavior-policy choices through objectives
that aggregate estimator variances across target policies.
They establish unbiasedness and give conditions on policy
similarity under which their estimators improve upon
separate on-policy evaluation.
Their implementation estimates the value and second-moment
quantities needed for behavior-policy construction from
offline transition and reward data, using fitted policy
evaluation.
They also demonstrate the approach in Gridworld and
MuJoCo environments.
Our objective instead controls the maximum uncertainty
across target policies, matching the requirement that
every value estimate satisfy the same accuracy guarantee.
Our planning procedure uses reward-free feature queries,
and our analysis accounts for the samples needed to learn
the design as part of a simultaneous high-probability
evaluation guarantee.

These works address sequential decision-making in MDPs,
where actions affect future state distributions.
We focus on stochastic contextual bandits with linear
conditional mean rewards and a known feature map.
This structure allows us to share information across
context--action pairs and obtain sample bounds governed
by feature dimension and the instance-dependent quantity
$\rho\opt$.
Our guarantees avoid explicit dependence on the
cardinalities of the context and action spaces and require
no numerical lower bound on second-moment eigenvalues.
The distinction between context draws, policy queries,
and reward observations is essential: the shared contexts
reduce the context budget, while querying every target
policy still requires $K$ policy queries per context.

\section{Planning phase}\label{app:planning phase}

We first recall Bernstein's inequality for bounded random variables,
which we use to prove \Cref{lem:V_hat_population}.

\begin{lemma}[Bernstein inequality
    {\cite[Theorem~2.8.4]{vershynin2019high}}]
\label{lem:bernstein}
    Let $Y_1,\ldots,Y_N$ be independent random variables satisfying
    $\E[Y_i]=0$ and $\abs{Y_i}\le b$ almost surely, for some $b>0$
    and every $i\in[N]$. Set
    $\sigma^2:=\sum_{i=1}^N\E[Y_i^2]$.
    For every $t>0$,
    \[
        \P\left(\abs{\sum_{i=1}^N Y_i}\ge t\right)
        \le 2\exp\left(
            -\frac{t^2}{2(\sigma^2+bt/3)}
        \right).
    \]
\end{lemma}

\subsection{\texorpdfstring{Proof of \Cref{lem:V_hat_population}}
    {Proof of the population value-estimation lemma}}
\label{app:proof_V_hat_population}

\csname Vhat_population\endcsname*

\begin{proof}
Fix a mixture $\bmu\in\triangle^{K-1}$ and a target policy
$\pi_k$, $k\in[K]$. The samples are independent, with
$X_t\sim\CP$, $I_t\sim\bmu$ independently of $X_t$, and
$A_t\sim\pi_{I_t}(\cdot\mid X_t)$.
Write
\[
    \phi_t:=\phi(X_t,A_t),\qquad
    w_k:=w_{\pi_k},\qquad
    V_k:=V_{\pi_k},\qquad
    C_R:=LS+\beta,\qquad
    q_k:=\bphi{k}^\top\Sigma_{\bmu}^{-1}\bphi{k}.
\]
Define the centered random variables
\[
    Z_t:=w_k(X_t,A_t)R_t-V_k,\qquad t\in[N].
\]
Then
$\hat{V}_{\pi_k}-V_k=N^{-1}\sum_{t=1}^N Z_t$.
Using the conditional mean-zero noise assumption, we have
\begin{align*}
    \E[Z_t]
    &= \E_{\bmu}[w_k(X_t,A_t)R_t]-V_k\\
    &= \E_{\bmu}\!\left[
         w_k(X_t,A_t)\dotp{\phi_t}{\theta\opt}
       \right]
       +\E_{\bmu}\!\left[
         w_k(X_t,A_t)\E[\eta_t\mid X_t,A_t]
       \right]-V_k\\
    &= \bphi{k}^\top\Sigma_{\bmu}^{-1}
       \E_{\bmu}[\phi_t\phi_t^\top]\theta\opt-V_k\\
    &= \bphi{k}^\top\Sigma_{\bmu}^{-1}
       \Sigma_{\bmu}\theta\opt-V_k\\
    &= \bphi{k}^\top\theta\opt-V_k\\
    &=0.
\end{align*}

Next, we bound the magnitude of $Z_t$.
By \Cref{assum:bounded_params} and the noise bound,
\begin{align*}
    \abs{R_t}
    &\le \abs{\dotp{\phi_t}{\theta\opt}}+\abs{\eta_t}\\
    &\le \|\phi_t\|_2\|\theta\opt\|_2+\beta\\
    &\le LS+\beta=C_R.
\end{align*}
Since $\E_{\bmu}[w_k(X_t,A_t)R_t]=V_k$, it follows that
\begin{align*}
    \abs{V_k}
    &\le \E_{\bmu}\!\left[\abs{w_k(X_t,A_t)R_t}\right]\\
    &\le C_R\,\E_{\bmu}\!\left[\abs{w_k(X_t,A_t)}\right]\\
    &\le C_R W_{\bmu}.
\end{align*}
Consequently,
\begin{align*}
    \abs{Z_t}
    &\le \abs{w_k(X_t,A_t)R_t}+\abs{V_k}\\
    &\le 2C_R W_{\bmu}.
\end{align*}

We now bound the variance. Since the weighted reward has mean $V_k$,
\begin{align*}
    \E[Z_t^2]
    &=\E_{\bmu}\!\left[
        \big(w_k(X_t,A_t)R_t-V_k\big)^2
      \right]\\
    &=\E_{\bmu}\!\left[w_k(X_t,A_t)^2R_t^2\right]-V_k^2\\
    &\le C_R^2\E_{\bmu}\!\left[w_k(X_t,A_t)^2\right]\\
    &=C_R^2\bphi{k}^\top\Sigma_{\bmu}^{-1}
      \E_{\bmu}[\phi_t\phi_t^\top]
      \Sigma_{\bmu}^{-1}\bphi{k}\\
    &=C_R^2\bphi{k}^\top\Sigma_{\bmu}^{-1}
      \Sigma_{\bmu}\Sigma_{\bmu}^{-1}\bphi{k}\\
    &=C_R^2q_k.
\end{align*}
Thus,
\[
    \sigma_k^2:=\sum_{t=1}^N\E[Z_t^2]\le NC_R^2q_k.
\]

If $W_{\bmu}=0$, all weights vanish almost surely and the assertion
follows immediately. Otherwise, set $b:=2C_R W_{\bmu}>0$.
Applying \Cref{lem:bernstein}, for every $t>0$ we obtain
\begin{align*}
    \P\left(\abs{\hat{V}_{\pi_k}-V_k}\ge t\right)
    &=\P\left(\abs{\sum_{s=1}^N Z_s}\ge Nt\right)\\
    &\le 2\exp\left(
       -\frac{N^2t^2}{2(\sigma_k^2+bNt/3)}
       \right)\\
    &\le 2\exp\left(
       -\frac{Nt^2}{2(C_R^2q_k+bt/3)}
       \right).
\end{align*}

To invert this bound, let $\ell:=\log(2K/\delta)$.
The positive root of
\[
    Nt^2-\frac{2b\ell}{3}t-2C_R^2q_k\ell=0
\]
is
\[
    t_k=\frac{b\ell}{3N}
        +\sqrt{\frac{2C_R^2q_k\ell}{N}
                    +\left(\frac{b\ell}{3N}\right)^2}.
\]
At $t=t_k$, the exponent equals $-\ell$.
Using $\sqrt{u+v}\le\sqrt{u}+\sqrt{v}$ for $u,v\ge0$ gives
\begin{align*}
    t_k
    &\le C_R\sqrt{\frac{2q_k\ell}{N}}
          +\frac{2b\ell}{3N}\\
    &=C_R\left(
         \sqrt{\frac{2q_k\ell}{N}}
         +\frac{4W_{\bmu}\ell}{3N}
       \right)\\
    &\le C_R\left(
         \sqrt{\frac{2\rho(\bmu)\ell}{N}}
         +\frac{4W_{\bmu}\ell}{3N}
       \right)
       =:\tau.
\end{align*}
The last inequality follows from $q_k\le\rho(\bmu)$.
A union bound over the $K$ target policies therefore yields
\begin{align*}
    \P\left(
        \max_{k\in[K]}\abs{\hat{V}_{\pi_k}-V_k}\ge\tau
    \right)
    &\le\sum_{k=1}^K
       \P\left(\abs{\hat{V}_{\pi_k}-V_k}\ge\tau\right)\\
    &\le\sum_{k=1}^K 2e^{-\ell}\\
    &=\delta.
\end{align*}
Substituting $C_R=LS+\beta$ and $\ell=\log(2K/\delta)$ proves
the stated bound.

If $\bmu$ is selected using independent planning data, condition
on those data. The mixture is then fixed, and the fresh evaluation
samples remain independent with the required distribution.
The same argument proves the conditional guarantee.
\end{proof}

\subsection{\texorpdfstring{Proof of \Cref{prop:bound_max_uncertainty}}
    {Proof of the design complexity bound}}
\label{app:proof_bound_max_uncertainty}

\maxUncertaintyBound*

\begin{proof}
For $\bmu,\alpha\in\triangle^{K-1}$, define
\[
    F(\bmu,\alpha)
    :=\sum_{k=1}^K\alpha_k
       \bphi{k}^\top\Sigma_{\bmu}^{-1}\bphi{k}.
\]
By \Cref{assum:coverage}, $F$ is continuous on the product
of the two simplices. It is affine in $\alpha$ and convex in $\bmu$.
Indeed,
\[
    \bphi{k}^\top\Sigma_{\bmu}^{-1}\bphi{k}
    =\max_{z\in\mathbb{R}^d}
      \left\{2\bphi{k}^\top z-z^\top\Sigma_{\bmu}z\right\},
\]
which is a pointwise maximum of affine functions of $\bmu$.

Both simplices are compact and convex, so Sion's minimax theorem
\citep{sion1958general} gives
\begin{align*}
    \rho\opt
    &=\min_{\bmu\in\triangle^{K-1}}
      \max_{\alpha\in\triangle^{K-1}}F(\bmu,\alpha)\\
    &=\max_{\alpha\in\triangle^{K-1}}
      \min_{\bmu\in\triangle^{K-1}}F(\bmu,\alpha)\\
    &\le\max_{\alpha\in\triangle^{K-1}}F(\alpha,\alpha).
\end{align*}
The first equality uses the fact that a linear function on the
simplex attains its maximum at a vertex. The last inequality
follows by choosing $\bmu=\alpha$ in the inner minimization.

For any $\alpha\in\triangle^{K-1}$, write
$\Sigma_\alpha:=\sum_{k=1}^K\alpha_k\Sigma_k$.
Jensen's inequality for the convex quadratic form
$v\mapsto v^\top\Sigma_\alpha^{-1}v$ yields
\begin{align*}
    F(\alpha,\alpha)
    &=\sum_{k=1}^K\alpha_k
      \big(\E_{\pi_k}\phi(X,A)\big)^\top
      \Sigma_\alpha^{-1}
      \big(\E_{\pi_k}\phi(X,A)\big)\\
    &\le\sum_{k=1}^K\alpha_k
      \E_{\pi_k}\!\left[
        \phi(X,A)^\top\Sigma_\alpha^{-1}\phi(X,A)
      \right]\\
    &=\operatorname{tr}\!\left(
        \Sigma_\alpha^{-1}
        \sum_{k=1}^K\alpha_k
        \E_{\pi_k}[\phi(X,A)\phi(X,A)^\top]
      \right)\\
    &=\operatorname{tr}(\Sigma_\alpha^{-1}\Sigma_\alpha)\\
    &=d.
\end{align*}
Combining the two displays proves $\rho\opt\le d$.
\end{proof}

\subsection{\texorpdfstring
    {Proof of \Cref{lem:hat_phi_estimation}}
    {Proof of the feature mean estimation lemma}}
\label{app:proof_hat_phi_estimation}

We next state the bounded differences inequality, which we use
to control the error in estimating the feature means.

\begin{lemma}[McDiarmid's inequality
    {\cite[Theorem~2.9.1]{vershynin2019high}}]
\label{lem:mcdiarmid}
    Let $Y_1,\ldots,Y_N$ be independent random variables
    taking values in a measurable space $\mathcal{Y}$.
    Suppose a measurable function
    $f:\mathcal{Y}^N \to \mathbb{R}$ satisfies
    \[
        \left|
        f(y_1,\ldots,y_i,\ldots,y_N)
        -
        f(y_1,\ldots,y_i',\ldots,y_N)
        \right|
        \le c_i
    \]
    for every $i \in [N]$ and all arguments in its domain.
    Then, for every $u>0$,
    \[
        \P\left(
        f(Y_1,\ldots,Y_N)-\E f(Y_1,\ldots,Y_N)
        \ge u
        \right)
        \le
        \exp\left(-\frac{2u^2}{\sum_{i=1}^N c_i^2}\right).
    \]
\end{lemma}

\hatPhiEstimation*

\begin{proof}
Fix a policy $k \in [K]$, and write
$Y_t := \phi(X_t,A_t^k)$ and $m_k := \bphi{k}$.
The vectors $Y_1,\ldots,Y_N$ are independent and identically
distributed, with $\E Y_t=m_k$ and $\norm{Y_t}_2 \le L$
almost surely.
Define
\[
    f_k(y_1,\ldots,y_N)
    :=
    \norm{\frac{1}{N}\sum_{t=1}^N y_t-m_k}_2,
\]
on the product of the Euclidean balls of radius $L$.
Thus,
$f_k(Y_1,\ldots,Y_N)=\norm{\hat{\phi}^k_N-\bphi{k}}_2$.

We first bound the expected estimation error.
By independence and centering, for $s \ne t$,
\[
    \E\left[(Y_s-m_k)^\top(Y_t-m_k)\right]
    =
    \left(\E[Y_s-m_k]\right)^\top
    \E[Y_t-m_k]
    =0.
\]
Consequently,
\begin{align*}
    \E\norm{\hat{\phi}^k_N-m_k}_2^2
    &=
    \frac{1}{N^2}
    \sum_{s=1}^N\sum_{t=1}^N
    \E\left[(Y_s-m_k)^\top(Y_t-m_k)\right] \\
    &=
    \frac{1}{N^2}
    \sum_{t=1}^N \E\norm{Y_t-m_k}_2^2 \\
    &=
    \frac{1}{N^2}
    \sum_{t=1}^N
    \left(\E\norm{Y_t}_2^2-\norm{m_k}_2^2\right) \\
    &\le \frac{L^2}{N}.
\end{align*}
Jensen's inequality therefore gives
\begin{align*}
    \E f_k(Y_1,\ldots,Y_N)
    &=
    \E\norm{\hat{\phi}^k_N-m_k}_2 \\
    &\le
    \sqrt{\E\norm{\hat{\phi}^k_N-m_k}_2^2}
    \le \frac{L}{\sqrt{N}}.
\end{align*}

We now verify the bounded differences condition.
If only the $i$th argument changes, the reverse triangle
inequality yields
\begin{align*}
    &\left|
    f_k(y_1,\ldots,y_i,\ldots,y_N)
    -
    f_k(y_1,\ldots,y_i',\ldots,y_N)
    \right| \\
    &\qquad\le
    \norm{\frac{y_i-y_i'}{N}}_2 \\
    &\qquad\le
    \frac{\norm{y_i}_2+\norm{y_i'}_2}{N}
    \le \frac{2L}{N}.
\end{align*}
For $L>0$, applying \Cref{lem:mcdiarmid} with
$c_i=2L/N$ gives, for every $u>0$,
\begin{align*}
    \P\left(
    \norm{\hat{\phi}^k_N-\bphi{k}}_2
    \ge \frac{L}{\sqrt{N}}+u
    \right)
    &\le
    \P\left(f_k-\E f_k \ge u\right) \\
    &\le
    \exp\left(-\frac{Nu^2}{2L^2}\right).
\end{align*}
Taking
\[
    u=L\sqrt{\frac{2\log(K/\delta)}{N}}
\]
bounds this probability by $\delta/K$.
A union bound over the $K$ target policies now yields
\begin{align*}
    &\P\left(
    \exists k \in [K]:
    \norm{\hat{\phi}^k_N-\bphi{k}}_2
    \ge
    \frac{L}{\sqrt{N}}
    \left(1+\sqrt{2\log\left(\frac{K}{\delta}\right)}\right)
    \right) \\
    &\qquad\le
    \sum_{k=1}^K \frac{\delta}{K}
    =\delta.
\end{align*}
\end{proof}

\subsection{\texorpdfstring
    {Proof of \Cref{lem:hatSigmaEstimation}}
    {Proof of the regularized second-moment estimation lemma}}
\label{app:proof_hatSigmaEstimation}

To prove the regularized second-moment bound, we use the
following matrix version of Bernstein's inequality.
Throughout this proof, $\|\cdot\|_{\mathrm{op}}$ denotes
the operator norm.

\begin{lemma}[Matrix Bernstein's inequality
    {\cite[Theorem~5.4.1]{vershynin2019high}}]
\label{lem:matrix_bernstein}
    Let $Y_1,\ldots,Y_N$ be independent symmetric
    $d \times d$ random matrices satisfying
    $\E Y_t=0$ and $\|Y_t\|_{\mathrm{op}}\le R$
    almost surely.
    Define
    \[
        v :=
        \left\|\sum_{t=1}^N \E[Y_t^2]\right\|_{\mathrm{op}}.
    \]
    Then, for every $s>0$,
    \[
        \P\left(
        \left\|\sum_{t=1}^N Y_t\right\|_{\mathrm{op}}
        \ge s
        \right)
        \le
        2d\exp\left(
        -\frac{s^2}{2(v+Rs/3)}
        \right).
    \]
\end{lemma}

\hatSigmaEstimation*

\begin{proof}
Fix a policy $k \in [K]$, and write
$U_t:=\phi(X_t,A_t^k)$ and $G:=\Sigma_k+\lambda I$.
The vectors $U_1,\ldots,U_N$ are independent and identically
distributed, with
$\E[U_tU_t^\top]=\Sigma_k$ and $\|U_t\|_2\le L$
almost surely.
Since $G \succeq \lambda I$, it is positive definite.

We apply matrix Bernstein to the normalized error.
Define
\[
    Z_t:=G^{-1/2}U_tU_t^\top G^{-1/2},
    \qquad
    M:=\E Z_t=G^{-1/2}\Sigma_kG^{-1/2},
    \qquad
    Y_t:=Z_t-M.
\]
The matrices $Y_t$ are independent, symmetric, and mean zero,
and
\[
    \frac1N\sum_{t=1}^N Y_t
    =
    G^{-1/2}(\hat{\Sigma}_k-\Sigma_k)G^{-1/2}.
\]
The population matrix $G$ is used only in the analysis.

We first bound the operator norm of each $Y_t$.
The feature bound implies
\[
    0 \preceq U_tU_t^\top \preceq L^2I,
    \qquad
    0 \preceq \Sigma_k \preceq L^2I.
\]
Consequently,
$-L^2I \preceq U_tU_t^\top-\Sigma_k \preceq L^2I$,
and hence
\begin{align*}
    \|Y_t\|_{\mathrm{op}}
    &=
    \left\|
    G^{-1/2}(U_tU_t^\top-\Sigma_k)G^{-1/2}
    \right\|_{\mathrm{op}} \\
    &\le
    \|G^{-1/2}\|_{\mathrm{op}}^2
    \|U_tU_t^\top-\Sigma_k\|_{\mathrm{op}} \\
    &\le \frac{L^2}{\lambda}.
\end{align*}

Next, we bound the matrix variance.
By congruence and $G \succeq \lambda I$,
\[
    0 \preceq Z_t
    \preceq L^2G^{-1}
    \preceq \frac{L^2}{\lambda}I,
    \qquad
    0 \preceq M
    = I-\lambda G^{-1}
    \preceq I.
\]
Every eigenvalue of $Z_t$ therefore belongs to
$[0,L^2/\lambda]$, which gives
\[
    Z_t^2 \preceq \frac{L^2}{\lambda}Z_t.
\]
Since $M$ is deterministic and $\E Z_t=M$,
\begin{align*}
    \E[Y_t^2]
    &=
    \E[(Z_t-M)^2] \\
    &=
    \E[Z_t^2]
    -\E[Z_t]M-M\E[Z_t]+M^2 \\
    &=
    \E[Z_t^2]-M^2 \\
    &\preceq
    \E[Z_t^2] \\
    &\preceq
    \frac{L^2}{\lambda}\E[Z_t] \\
    &=
    \frac{L^2}{\lambda}M
    \preceq
    \frac{L^2}{\lambda}I.
\end{align*}
Thus,
\[
    v:=
    \left\|\sum_{t=1}^N\E[Y_t^2]\right\|_{\mathrm{op}}
    \le \frac{NL^2}{\lambda}.
\]

Assume $L>0$; the case $L=0$ is immediate.
Applying \Cref{lem:matrix_bernstein} with
$R=L^2/\lambda$ and $s=N/2$ yields
\begin{align*}
    &\P\left(
    \left\|
    G^{-1/2}(\hat{\Sigma}_k-\Sigma_k)G^{-1/2}
    \right\|_{\mathrm{op}}
    \ge \frac12
    \right) \\
    &\qquad=
    \P\left(
    \left\|\sum_{t=1}^N Y_t\right\|_{\mathrm{op}}
    \ge \frac{N}{2}
    \right) \\
    &\qquad\le
    2d\exp\left(
    -\frac{N^2/4}{
    2\left(NL^2/\lambda+NL^2/(6\lambda)\right)}
    \right) \\
    &\qquad=
    2d\exp\left(-\frac{3N\lambda}{28L^2}\right) \\
    &\qquad\le \frac{\delta}{K}.
\end{align*}

On the complementary event, symmetry gives
\[
    -\frac12 I
    \preceq
    G^{-1/2}(\hat{\Sigma}_k-\Sigma_k)G^{-1/2}
    \preceq
    \frac12 I.
\]
Multiplying on both sides by $G^{1/2}$ and then adding $G$
gives
\begin{align*}
    -\frac12G
    &\preceq
    \hat{\Sigma}_k-\Sigma_k
    \preceq
    \frac12G, \\
    \frac12G
    &\preceq
    \hat{\Sigma}_k+\lambda I
    \preceq
    \frac32G.
\end{align*}
Substituting $G=\Sigma_k+\lambda I$ proves the desired
inequalities for policy $k$.
A union bound over $k \in [K]$ bounds the probability that
any of these inequalities fails by
$K(\delta/K)=\delta$.
Independence between policies sharing the same context
is not required.
\end{proof}

\subsection{\texorpdfstring
    {Proof of \Cref{lem:sampleOptProblem}}
    {Proof of the empirical design guarantee}}
\label{app:proof_sampleOptProblem}

\sampleOptProblem*

\begin{proof}
Write
\[
    \ell:=\log\left(\frac{4Kd}{\delta}\right).
\]
Apply \Cref{lem:hat_phi_estimation} with failure probability
$\delta/2$.
Since $\ell>1$ and $\log(2K/\delta)\le\ell$, the assumed
sample size gives, simultaneously for all $k \in [K]$,
\begin{align*}
    \|\hat{\phi}_k-\bphi{k}\|_2
    &\le
    \frac{L}{\sqrt{N}}
    \left(1+\sqrt{2\log\left(\frac{2K}{\delta}\right)}\right) \\
    &\le
    \frac{L}{\sqrt{N}}(1+\sqrt{2\ell}) \\
    &\le
    4L\sqrt{\frac{\ell}{N}}
    \le \sqrt{\lambda}.
\end{align*}
Also, $16\ge 28/3$, so
\Cref{lem:hatSigmaEstimation}, applied with failure
probability $\delta/2$, gives
\[
    \frac12(\Sigma_k+\lambda I)
    \preceq
    \hat{\Sigma}_k+\lambda I
    \preceq
    \frac32(\Sigma_k+\lambda I),
    \qquad k \in [K].
\]
By a union bound, both conclusions hold on an event
$\mathcal{E}$ of probability at least $1-\delta$.
The remainder of the proof is deterministic on this event.

For $\bmu \in \triangle^{K-1}$, define
\[
    G_{\bmu}:=\sum_{i=1}^K\mu_i\Sigma_i+\lambda I,
    \qquad
    \hat{G}_{\bmu}:=
    \sum_{i=1}^K\mu_i\hat{\Sigma}_i+\lambda I.
\]
Multiplying the second-moment inequalities by $\mu_i$
and summing over $i$, using $\sum_i\mu_i=1$, yields
\[
    \frac12G_{\bmu}
    \preceq \hat{G}_{\bmu}
    \preceq \frac32G_{\bmu}.
\]
These inequalities hold simultaneously for every mixture.
Inverting them gives
\[
    \hat{G}_{\bmu}^{-1}\preceq 2G_{\bmu}^{-1},
    \qquad
    G_{\bmu}^{-1}\preceq \frac32\hat{G}_{\bmu}^{-1}.
\]

Let $e_k:=\hat{\phi}_k-\bphi{k}$.
Since $G_{\bmu}\succeq\lambda I$, on $\mathcal{E}$,
\begin{align*}
    e_k^\top G_{\bmu}^{-1}e_k
    &\le \frac{\|e_k\|_2^2}{\lambda}
    \le 1.
\end{align*}
We will also use the elementary inequality
\[
    (a+b)^\top H(a+b)
    \le 2a^\top Ha+2b^\top Hb,
    \qquad H\succeq0,
\]
which follows by expanding $(a-b)^\top H(a-b)\ge0$.

We first compare the empirical objective with its population
counterpart.
For every policy $k$ and mixture $\bmu$,
\begin{align*}
    \hat{\phi}_k^\top\hat{G}_{\bmu}^{-1}\hat{\phi}_k
    &\le
    2\hat{\phi}_k^\top G_{\bmu}^{-1}\hat{\phi}_k \\
    &=
    2(\bphi{k}+e_k)^\top
    G_{\bmu}^{-1}(\bphi{k}+e_k) \\
    &\le
    4\bphi{k}^\top G_{\bmu}^{-1}\bphi{k}
    +4e_k^\top G_{\bmu}^{-1}e_k \\
    &\le
    4\bphi{k}^\top G_{\bmu}^{-1}\bphi{k}+4.
\end{align*}
Taking the maximum over $k$ gives
\[
    \hat{F}_\lambda(\bmu)\le 4F_\lambda(\bmu)+4.
\]

For the reverse comparison,
\begin{align*}
    \bphi{k}^\top G_{\bmu}^{-1}\bphi{k}
    &=
    (\hat{\phi}_k-e_k)^\top
    G_{\bmu}^{-1}(\hat{\phi}_k-e_k) \\
    &\le
    2\hat{\phi}_k^\top G_{\bmu}^{-1}\hat{\phi}_k
    +2e_k^\top G_{\bmu}^{-1}e_k \\
    &\le
    3\hat{\phi}_k^\top\hat{G}_{\bmu}^{-1}\hat{\phi}_k+2.
\end{align*}
Thus,
\[
    F_\lambda(\bmu)\le 3\hat{F}_\lambda(\bmu)+2.
\]
Both objective comparisons hold simultaneously for every
$\bmu \in \triangle^{K-1}$.

Finally, define
\[
    \rho_\lambda:=
    \min_{\bmu\in\triangle^{K-1}}F_\lambda(\bmu).
\]
Since
$(\Sigma_{\bmu}+\lambda I)^{-1}\preceq\Sigma_{\bmu}^{-1}$,
we have $\rho_\lambda\le\rho\opt$.
By optimality of $\hat{\bmu}$,
\begin{align*}
    \hat{F}_\lambda(\hat{\bmu})
    &=
    \min_{\bmu\in\triangle^{K-1}}\hat{F}_\lambda(\bmu) \\
    &\le
    \min_{\bmu\in\triangle^{K-1}}
    \bigl(4F_\lambda(\bmu)+4\bigr) \\
    &=4\rho_\lambda+4 \\
    &\le4\rho\opt+4.
\end{align*}
Applying the reverse comparison at $\hat{\bmu}$ now gives
\begin{align*}
    F_\lambda(\hat{\bmu})
    &\le 3\hat{F}_\lambda(\hat{\bmu})+2 \\
    &\le 12\rho_\lambda+14 \\
    &\le 12\rho\opt+14.
\end{align*}
This proves both claims.
\end{proof}

\section{Sampling phase}\label{app:sampling phase}

\subsection{\texorpdfstring
    {Proof of \Cref{lem:thetaError}}
    {Proof of the ridge estimation error identity}}
\label{app:proof_thetaError}

\thetaError*

\begin{proof}
Write $\phi_i:=\phi(X_i,A_i)$.
Substituting the reward model
$R_i=\phi_i^\top\theta\opt+\eta_i$ into $\tilde{g}$ gives
\begin{align*}
    \tilde{g}
    &=
    \frac{1}{M}\sum_{i=1}^M\phi_iR_i \\
    &=
    \frac{1}{M}\sum_{i=1}^M
    \phi_i\phi_i^\top\theta\opt
    +
    \frac{1}{M}\sum_{i=1}^M\eta_i\phi_i \\
    &=
    \tilde{\Sigma}\theta\opt
    +
    \frac{1}{M}\sum_{i=1}^M\eta_i\phi_i \\
    &=
    (B-\lambda I)\theta\opt
    +
    \frac{1}{M}\sum_{i=1}^M\eta_i\phi_i.
\end{align*}
Since $B\succeq\lambda I\succ0$, multiplying by $B^{-1}$
and subtracting $\theta\opt$ yields
\begin{align*}
    \hat{\theta}-\theta\opt
    &=
    B^{-1}(B-\lambda I)\theta\opt-\theta\opt
    +
    \frac{1}{M}B^{-1}\sum_{i=1}^M\eta_i\phi_i \\
    &=
    -\lambda B^{-1}\theta\opt
    +
    \frac{1}{M}B^{-1}\sum_{i=1}^M\eta_i\phi_i.
\end{align*}
\end{proof}

We use the following two-sided consequence of Hoeffding's
inequality.

\begin{lemma}[Hoeffding's inequality
    {\cite[Theorem~2.2.6]{vershynin2019high}}]
\label{lem:hoeffding_bounded}
    Let $Y_1,\ldots,Y_M$ be independent random variables
    satisfying $a_i\le Y_i\le b_i$ almost surely.
    Then, for every $t>0$,
    \[
        \P\left(
        \left|\sum_{i=1}^M(Y_i-\E Y_i)\right|\ge t
        \right)
        \le
        2\exp\left(
        -\frac{2t^2}{\sum_{i=1}^M(b_i-a_i)^2}
        \right).
    \]
\end{lemma}

\subsection{\texorpdfstring
    {Proof of \Cref{lem:boundNoiseTerm}}
    {Proof of the reward-noise bound}}
\label{app:proof_boundNoiseTerm}

\boundNoiseTerm*

\begin{proof}
Let $\mathcal{D}$ denote the sigma-field generated by the
planning data, the independent feature-mean batch, and
the contexts and actions in the reward batch.
Conditional on $\mathcal{D}$, the matrices $B$,
the vectors $\tilde{\phi}_k$, and the features
$\phi_i:=\phi(X_i,A_i)$ are fixed.
Moreover, the noises are conditionally independent and satisfy
\[
    \E[\eta_i\mid\mathcal{D}]=0,
    \qquad
    |\eta_i|\le\beta.
\]

Fix $k\in[K]$ and define
\[
    c_{i,k}:=
    \frac{1}{M}\tilde{\phi}_k^\top B^{-1}\phi_i,
    \qquad
    T_k:=\sum_{i=1}^M c_{i,k}\eta_i.
\]
We first control the squared coefficients:
\begin{align*}
    \sum_{i=1}^M c_{i,k}^2
    &=
    \frac{1}{M^2}
    \sum_{i=1}^M
    \tilde{\phi}_k^\top
    B^{-1}\phi_i\phi_i^\top B^{-1}
    \tilde{\phi}_k \\
    &=
    \frac{1}{M}
    \tilde{\phi}_k^\top
    B^{-1}\tilde{\Sigma}B^{-1}
    \tilde{\phi}_k \\
    &=
    \frac{1}{M}
    \tilde{\phi}_k^\top
    B^{-1}(B-\lambda I)B^{-1}
    \tilde{\phi}_k \\
    &=
    \frac{1}{M}
    \left(
        Q_k-\lambda\tilde{\phi}_k^\top B^{-2}
        \tilde{\phi}_k
    \right) \\
    &\le \frac{Q_k}{M}.
\end{align*}

Conditional on $\mathcal{D}$, each $c_{i,k}\eta_i$
has mean zero and lies in
$[-\beta|c_{i,k}|,\beta|c_{i,k}|]$.
Therefore, \Cref{lem:hoeffding_bounded} gives
\begin{align*}
    \P\left(|T_k|\ge t\mid\mathcal{D}\right)
    &\le
    2\exp\left(
        -\frac{2t^2}{
        \sum_{i=1}^M(2\beta|c_{i,k}|)^2}
    \right) \\
    &=
    2\exp\left(
        -\frac{t^2}{
        2\beta^2\sum_{i=1}^M c_{i,k}^2}
    \right) \\
    &\le
    2\exp\left(-\frac{Mt^2}{2\beta^2Q_k}\right).
\end{align*}
These expressions apply when the denominators are positive.
If $\beta=0$ or $\sum_i c_{i,k}^2=0$, then $T_k=0$
almost surely and the desired conclusion is immediate.

For $Q_k>0$ and $\beta>0$, take
\[
    t_k:=
    \beta\sqrt{\frac{2Q_k\log(2K/\zeta)}{M}}.
\]
The conditional failure probability is at most $\zeta/K$.
A union bound over the target policies consequently gives
\[
    \P\left(
        \exists k\in[K]:
        |T_k|>
        \beta\sqrt{\frac{2Q_k\log(2K/\zeta)}{M}}
        \,\middle|\,\mathcal{D}
    \right)
    \le\zeta.
\]
Finally, taking expectations over $\mathcal{D}$ proves
the unconditional guarantee.
\end{proof}

\section{Algorithm and guarantees}\label{app:algorithm}

\subsection{\texorpdfstring
    {Proof of \Cref{thm:main_guarantee}}
    {Proof of the main guarantee}}
\label{app:proof_main_guarantee}

\mainGuarantee*

\begin{proof}
We first establish termination for every realization,
then define the concentration event used to prove accuracy
and the instance-dependent resource bounds.

For any planning batch, the empirical moments satisfy
\[
    \hat{\phi}_{j,k}\hat{\phi}_{j,k}^\top
    \preceq\hat{\Sigma}_{j,k},
    \qquad k\in[K].
\]
This follows because their difference is the empirical
covariance matrix.
Fix the batch and write
\[
    H_{\bmu}:=
    \sum_{i=1}^K\mu_i\hat{\Sigma}_{j,i}+\lambda_j I.
\]
Let $\bmu_D$ maximize $\log\det H_{\bmu}$ over the simplex.
A maximizer exists because $H_{\bmu}\succeq\lambda_j I$
and the simplex is compact.
The one-sided derivative towards the $k$th simplex vertex
is nonpositive, so
\[
    \operatorname{tr}\left(
        H_{\bmu_D}^{-1}
        \left(
            \hat{\Sigma}_{j,k}
            -\sum_{i=1}^K\mu_{D,i}\hat{\Sigma}_{j,i}
        \right)
    \right)
    \le0.
\]
Consequently,
\begin{align*}
    \hat{\phi}_{j,k}^\top H_{\bmu_D}^{-1}\hat{\phi}_{j,k}
    &\le
    \operatorname{tr}\left(
        H_{\bmu_D}^{-1}\hat{\Sigma}_{j,k}
    \right)\\
    &\le
    \operatorname{tr}\left(
        H_{\bmu_D}^{-1}(H_{\bmu_D}-\lambda_j I)
    \right)\\
    &=d-\lambda_j\operatorname{tr}(H_{\bmu_D}^{-1})
    \le d.
\end{align*}
Thus $\min_{\bmu}\hat{F}_{\lambda_j}(\bmu)\le d$
for every batch.
Since $r_J\ge d$, stage $J$ necessarily satisfies the
acceptance criterion.

For every executed planning stage $j$, define
$\mathcal{E}_{\mathrm{plan},j}$ by the simultaneous conditions
\[
    \|\hat{\phi}_{j,k}-\bphi{k}\|_2\le\sqrt{\lambda_j},
    \qquad
    \frac12(\Sigma_k+\lambda_j I)
    \preceq\hat{\Sigma}_{j,k}+\lambda_j I
    \preceq\frac32(\Sigma_k+\lambda_j I),
    \quad k\in[K].
\]
For an unexecuted stage, take
$\mathcal{E}_{\mathrm{plan},j}$ to be the whole sample space.
The argument in \Cref{lem:sampleOptProblem}, applied with
failure probability $\zeta$, shows that each executed
stage fails these conditions with conditional probability
at most $\zeta$.

For the final batches, define
\begin{align*}
    \mathcal{E}_{\mathrm{mean}}
    &:=
    \left\{
        \|\tilde{\phi}_k-\bphi{k}\|_2\le\sqrt{\lambda}
        \text{ for every }k\in[K]
    \right\},\\
    \mathcal{E}_{\mathrm{cov}}
    &:=
    \left\{
        \frac12G\preceq B\preceq\frac32G
    \right\},
    \qquad G:=\Sigma_{\hat{\bmu}}+\lambda I,\\
    \mathcal{E}_{\mathrm{noise}}
    &:=
    \left\{
        |T_k|
        \le
        \beta\sqrt{\frac{2Q_k\log(2K/\zeta)}{M}}
        \text{ for every }k\in[K]
    \right\},
\end{align*}
where
\[
    T_k:=
    \frac1M\tilde{\phi}_k^\top B^{-1}
    \sum_{i=1}^M\eta_i\phi(X_i,A_i).
\]
The sample budgets and
\Cref{lem:hat_phi_estimation,lem:hatSigmaEstimation,lem:boundNoiseTerm}
give failure probability at most $\zeta$ for each final event,
conditional on the data used to select the stage.
For the covariance event, we apply the concentration argument
to the single fixed mixture after conditioning on planning.

Define
\[
    \mathcal{E}
    :=
    \left(\bigcap_{j=0}^J\mathcal{E}_{\mathrm{plan},j}\right)
    \cap\mathcal{E}_{\mathrm{mean}}
    \cap\mathcal{E}_{\mathrm{cov}}
    \cap\mathcal{E}_{\mathrm{noise}}.
\]
A union bound, together with the conditional guarantees,
gives
\[
    \P(\mathcal{E}^c)
    \le(J+1)\zeta+3\zeta
    =\delta.
\]

We now work on $\mathcal{E}$.
Let $j^\star$ be the first stage with
$r_{j^\star}\ge1\vee\rho\opt$.
Such a stage exists because $\rho\opt\le d$.
At this stage,
\[
    \hat{F}_{\lambda_{j^\star}}(\hat{\bmu}_{j^\star})
    \le4\rho\opt+4
    \le8r_{j^\star},
\]
so the algorithm accepts no later than $j^\star$.
Therefore,
\[
    r_{\hat{j}}\le r_{j^\star}
    \le2(1\vee\rho\opt).
\]

Write $r:=r_{\hat{j}}$.
The acceptance rule and the objective comparisons in
\Cref{lem:sampleOptProblem} yield $Q_k\le120r$.
Using \Cref{eq:valueDecomp}, we obtain
\begin{align*}
    |V_k-\hat{V}_k|
    &\le
    S\sqrt{\lambda}
    +S\sqrt{\lambda Q_k}
    +|T_k|\\
    &\le
    \frac{\varepsilon}{\sqrt{2048r}}
    +\varepsilon\sqrt{\frac{120}{2048}}
    +\beta\sqrt{\frac{240r\ell}{M}}\\
    &\le
    \frac{\varepsilon}{2}
    +\frac{\varepsilon}{4}
    +\frac{\varepsilon}{4}
    =\varepsilon.
\end{align*}
For the final term, if $\beta>0$, use
$M\ge4096\beta^2r\ell/\varepsilon^2$ and
$240/4096\le1/16$.
If $\beta=0$, that term vanishes.

It remains to count resources.
Set
\[
    a:=\frac{32768L^2S^2\ell}{\varepsilon^2},
    \qquad
    b:=\frac{4096\beta^2\ell}{\varepsilon^2}.
\]
Then
$n_j=\lceil ar_j\rceil$ and
$M_j=\lceil\max\{a,b\}r_j\rceil$.
Since $\varepsilon<LS$, we have $a\ge1$.
Using $\sum_{j=0}^{\hat{j}}2^j<2r$,
\begin{align*}
    P:=\sum_{j=0}^{\hat{j}}n_j
    &\le2a\sum_{j=0}^{\hat{j}}r_j
    \le4ar,\\
    n_{\hat{j}}&\le2ar,\\
    M&\le2(a+b)r.
\end{align*}
The exact resource counts are
\[
    N_{\mathrm{contexts}}=P+n_{\hat{j}}+M,
    \qquad
    N_{\mathrm{rewards}}=M,
    \qquad
    N_{\mathrm{queries}}=K(P+n_{\hat{j}})+M.
\]
Substituting the preceding bounds and
$r\le2(1\vee\rho\opt)$ proves the asserted rates.
Finally, deterministic termination gives
$r\le2^J\le2d$ for every realization, which yields the
deterministic resource bounds.
\end{proof}

\end{document}